\documentclass[sigconf]{acmart}

\usepackage{multirow}
\usepackage{subcaption}
\usepackage[ruled,vlined]{algorithm2e}

\AtBeginDocument{%
  }

\setcopyright{acmcopyright}
\copyrightyear{2023}
\acmYear{2023}
\acmDOI{XXXXXXX.XXXXXXX}

\acmConference[ICMI '23]{International Conference on Multimodal Interaction}{Oct 09--13, 2023}{Paris, France}

\acmPrice{15.00}
\acmISBN{978-1-4503-XXXX-X/18/06}

\copyrightyear{2023} 
\acmYear{2023} 
\setcopyright{acmlicensed}\acmConference[ICMI '23]{INTERNATIONAL CONFERENCE ON MULTIMODAL INTERACTION}{October 9--13, 2023}{Paris, France}
\acmBooktitle{INTERNATIONAL CONFERENCE ON MULTIMODAL INTERACTION (ICMI '23), October 9--13, 2023, Paris, France}
\acmPrice{15.00}
\acmDOI{10.1145/3577190.3614174}
\acmISBN{979-8-4007-0055-2/23/10}

\begin{document}

\title{Recognizing Intent in Collaborative Manipulation}




\author{Zhanibek Rysbek}
\authornote{Both authors contributed equally to this research.}
\affiliation{%
  \institution{University of Illinois Chicago}
  \city{Chicago}
  \state{IL}
  \country{USA}
  \postcode{43017-6221}
}
\email{zrysbe2@uic.edu}

\author{Ki Hwan Oh}
\authornotemark[1]
\affiliation{%
  \institution{University of Illinois Chicago}
  \city{Chicago} 
  \state{IL}
  \country{USA}}
\email{koh43@uic.edu}

\author{Milo\v s \v Zefran}
\affiliation{%
  \institution{University of Illinois Chicago}
  \city{Chicago}
  \state{IL}
  \country{USA}}
\email{mzefran@uic.edu}


\begin{abstract}
Collaborative manipulation is inherently multimodal, with haptic communication playing a central role. When performed by humans, it involves back-and-forth force exchanges between the participants through which they resolve possible conflicts and determine their roles. Much of the existing work on collaborative human-robot manipulation assumes that the robot follows the human. But for a robot to match the performance of a human partner it needs to be able to take initiative and lead when appropriate. To achieve such human-like performance, the robot needs to have the ability to (1) determine the intent of the human, (2) clearly express its own intent, and (3) choose its actions so that the dyad reaches consensus. This work proposes a framework for recognizing human intent in collaborative manipulation tasks using force exchanges. Grounded in a dataset collected during a human study, we introduce a set of features that can be computed from the measured signals and report the results of a classifier trained on our collected human-human interaction data. Two metrics are used to evaluate the intent recognizer: overall accuracy and the ability to correctly identify transitions. The proposed recognizer shows robustness against the variations in the partner's actions and the confounding effects due to the variability in grasp forces and dynamic effects of walking. The results demonstrate that the proposed recognizer is well-suited for implementation in a physical interaction control scheme.

\end{abstract}





\maketitle

\section{Introduction}

\begin{figure*}[t]
    \centering
    \includegraphics[width=\textwidth, trim={2cm 3.7cm 1.8cm 4.4cm}, clip]{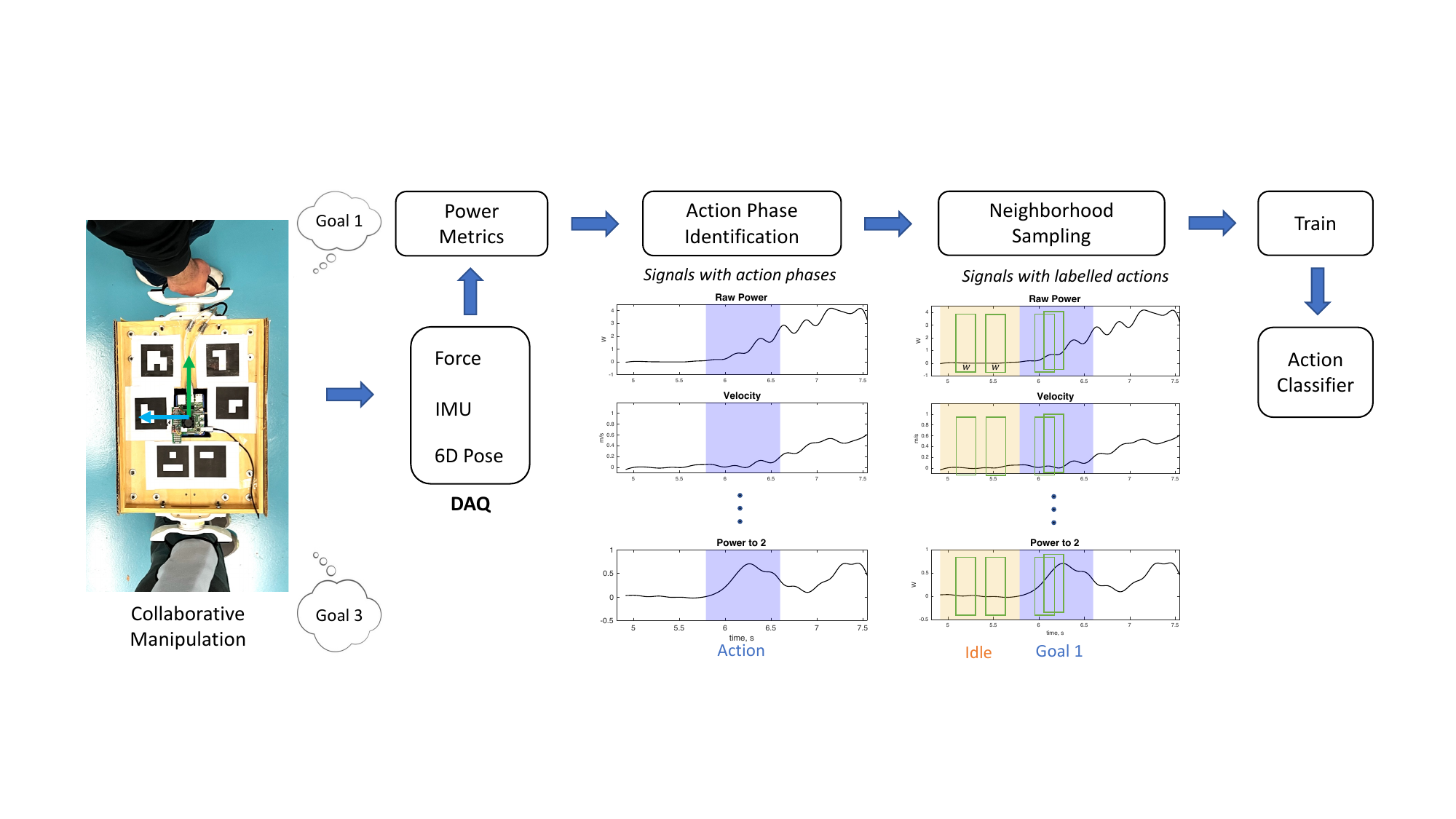}
    \caption{Training pipeline. Force-kinematics data is generated from co-manipulation experiments. Then, action phases are identified based on the power features. Then training data is generated by neighborhood sampling.}
    \label{fig:training_scheme}
\end{figure*}



In daily life, humans communicate across many modalities simultaneously. In order to use robots actively in daily life, it is important to build an ability for the robot to engage in multi-modal interaction. In collaborative manipulation, humans employ speech and body pose as needed, but the haptic channel is active throughout the interaction. This work focuses on an underexplored area of haptic communication centered around a co-manipulation task. 

Recent work in physical Human-Robot Interaction (pHRI) uses the interaction between humans to inform the design of collaborative robots. Much of the existing work on collaborative human-robot manipulation assumes that the robot follows the human. But for a robot to match the performance of a human partner it needs to be able to take initiative and lead when appropriate. In particular, the robot must have the ability to (1) understand human intent, (2) express its own intent and (3) choose an action policy to reach an agreement. Towards that goal, this work proposes a framework for recognizing human intent in collaborative manipulation tasks using force exchanges.



Due to the difficulty of obtaining data where human intent can be clearly identified, no direct study addresses individual human action identification over a haptic medium in real-time. In the absence of direct supervision, authors in \cite{aiyama1999cooperative} suggested a collaborative object manipulation system with implicit communication based on the participant's signals. However, studies exist that predict human-human interaction patterns which are caused by individual dyad actions. For example, authors in \cite{madan_recognition} suggested a taxonomy of interaction patterns, and trained a classifier to predict them. Building on that work, \cite{alsaadi_resolving_conflicts} proposed a pHRI system, which switches to active collaboration mode when harmonious interaction is detected. Dynamic role allocation during natural human interaction has been studied in \cite{mortl_role_2012, noohi_Model, monaikul_role_2020, chen2015RolesofHOAct}, where dyads often switch to take the lead when required during the task. A body of research work is focused on studying human behaviors in haptic-related tasks. For example, \cite{mojtahediCommunicationInferenceIntended2017} conducted a study of how humans infer the directional intent of the partner with visual feedback. The conclusion was that humans adapt to partner behavior, and the accuracy of inference increases significantly after a short practice. In \cite{george_guiding_human_follower_2022}, authors trained Multi-layer Perceptron (MLP) to map force signals to velocity commands in overground guiding tasks. In follow-up work, they observed that humans learn impedance modulation strategies that increased the efficiency of haptic communication \cite{regmi2022humans}. The decision-making behavior of humans was studied in \cite{lokesh2022humans}, where researchers observed that humans spent a considerable amount of time gathering signal evidence from their partner's actions in a competing game experiment. 

When it comes to methodology in mining time series data for intent recognition, a common approach is to employ the sliding window technique \cite{varol_multiclas_realtime_intent} with clearly labeled regions. In the absence of supervision, authors in \cite{sana_unsup_representation_21} suggested a technique to separate window samples by discriminating out-of-neighborhood samples, given that the dataset regions can be coarsely separated by simple statistical methods.

In this work, we propose a framework for recognizing intent about the movement direction in collaborative manipulation. The main challenge of this approach is the difficulty of obtaining natural human action data that can be labeled for intent. To tackle this, we conducted a human study described in our previous work \cite{rysbek_physical_action2021,rysbek_robots_23}. In the experiment, participants individually received the intended goal configuration as well as the importance of reaching that configuration. In order to study actions that humans use to communicate their intent we implemented an action phase recognition algorithm described in Section \ref{sec:action_phase_detection}. By random sampling from the neighborhoods within the action phase, we generated a dataset that was subsequently used to train a classifier that can be used for intent recognition. We report the results of SVM, AdaBoost, and Fully connected Neural Networks classifiers. Along with the overall accuracy of these models, we report their transition accuracy after a voting filter is employed. We also provide a detailed discussion of the robustness of the classifier.

\vspace{-0.2cm}

\section{Human Action types}

\subsection{Definition}
In collaborative manipulation, dyads negotiate implicit and explicit task parameters through force exchanges \cite{gildert2018need}. Implicit parameters include grasping force, height of the movement plane, and speed of movement. They are typically determined through low-level coordination.
The most important parameter that is determined through explicit negotiation is the direction of movement. Conflict is common in this process and a difference in the intended direction of movement results in elevated values of the interaction force~\cite{rysbek_physical_action2021}. Several studies explicitly design the system to avoid such conflicts \cite{alsaadi_resolving_conflicts, groten_haptic_dominance}. For example, Participant 1 may wish to go towards a goal $g_i$ while Participant 2 tries to go to a different goal $g_j$. As a result, the manipulated object gets stretched (or compressed) and the magnitude of the stretch (compression) force (differences of the force signals) sharply increases.

When two humans collaborate in an environment, there are typically a finite number of locations that may be of interest during a particular activity. We will therefore assume that there are $N$ discrete goals $g_i, i \in \{1, \ldots, N\}$ in an environment and define a human action to be the intent of the participant to go to one of these predefined goal locations. We further consider the period before any movement takes place as an additional \textit{idle} action. Hence, a total number of classes is $N+1$, with the action set equal to $A=\{idle,g_1, g_2, ... , g_N\}$.



\subsection{Action Phase detection}
\label{sec:action_phase_detection}

In physical interaction, humans exchange push-pull actions to express their intent and probe the reaction of their partner. Such exchanges are critical in reaching an agreement on the final goal direction. Usually, their duration is brief compared to the duration of the entire task that contains lifting, negotiation, navigation, and release of the object. In order to study human-human interaction at the individual level, it is important to accurately detect the period when force exchanges actively convey intent as they shape the evolution of the interaction. We call this period the action phase.

Several studies~\cite{madan_recognition, rysbek_robots_23, rysbek_physical_action2021, noohi_quantitative_2014-2, alsaadi_resolving_conflicts, mortl_role_2012} observed that when push-pull actions are applied by human dyads in conflicting directions, force signals exhibit a distinctive hump-like pattern. Although these patterns consistently appear, relying solely on force signals does not sufficiently convey the cause of the conflict and the intention of each participant. Furthermore, when intent expression occurs without conflict, the characteristic hump-like patterns are absent, making the detection of the action phase and determination of the correct interaction state more intricate. To tackle this issue, we introduce power indices and put forth an algorithm for identifying the action phase, both outlined in the following sections.



\subsubsection{Power Indices}

Several studies use interaction power to distinguish different patterns of interaction \cite{madan_recognition, rysbek_robots_23}, often calculating averages over a period of time. In this study, we employ instantaneous interaction power, as introduced in \cite{rysbek_robots_23}:
\begin{equation}
    P_{k}(t) = F_{k}(t)\cdot v_{k}(t) =  \|F_{k}(t)\|\|v_{k}(t)\|\cos\left(\angle\left(F_{k}(t),v_{k}(t)\right)\right)
    \label{eq:power}
\end{equation}
where, $k \in \{1,2\}$ is the participant number, and $F_k$ and $v_k$ are force and velocity sensed at the grasp point, respectively. This metric provides insights into the overall direction of interaction. In the presence of $N$ goals, interaction power can be projected to the line that connects the current grasp point and goal:
\begin{equation}
\begin{split}
    P^{i}_{k} &= F^{i}_{k}\cdot v^{i}_{k} \\
    &= \|F_{k}\|\|v_{k}\|\cos(\angle(F_{k},g_i))\cos(\angle(v_{k},g_i))
\end{split}
\label{eq:power_proj}
\end{equation}
where $F^{i}_{k}$ and $v^{i}_{k}$ are projections of the force and velocity to a goal direction $g_i$. In contrast to general power in Equation \ref{eq:power}, projected power is the amount of total power dedicated to a specific direction that provides orientational information of the push-pull action.

In a noise-free environment, such signals alone are sufficient to detect the accurate intent of the participant by finding the maximum $P^{i}_k$ over possible goals $i$. Yet, due to the confounding components of the human force such as the grasp force, and force artifacts due to the dynamics of walking and maneuvering, the maximum power signal does not necessarily coincide with the direction truly intended by the participant. Therefore, we employ a supervised machine learning method that "learns" to eliminate these confounding factors, rendering it apt for real-time intent recognition.





\subsubsection{Algorithm}
\label{sec:first_mover}


\begin{figure}
    \centering
    \includegraphics[width=\columnwidth, trim={1.5cm 8.4cm 1.5cm 8.5cm},clip]{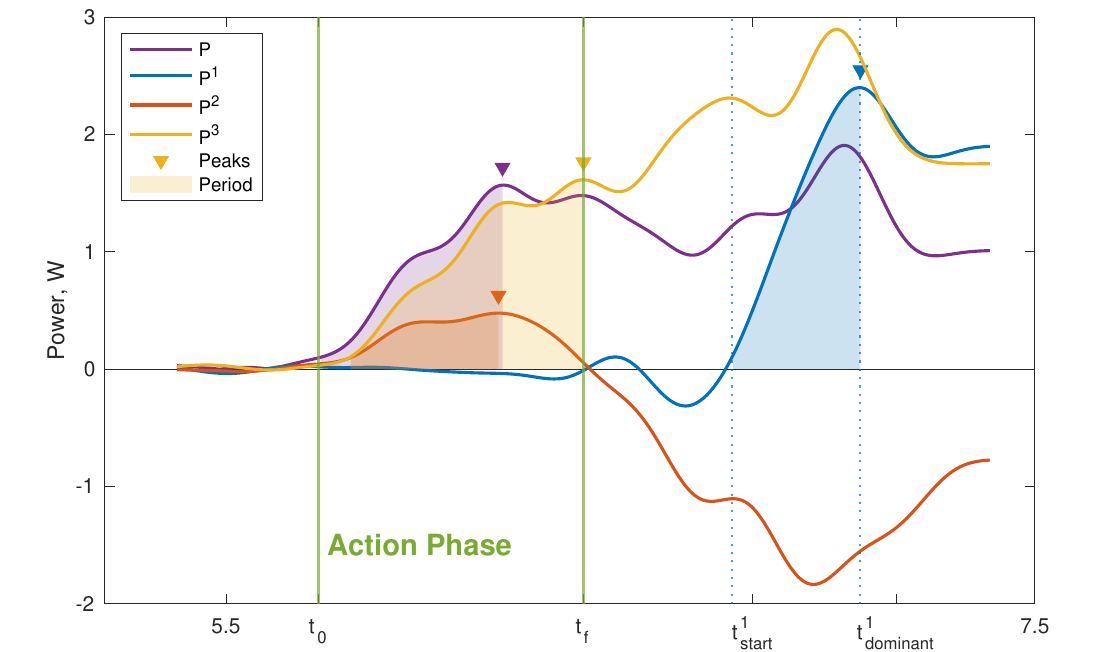}
    \caption{An example of the output of the Action Phase Detection Algorithm on Power signals.}
    \label{fig:action_phase}
\vspace{-0.5cm}
\end{figure}

In this section, we provide the details of how the action phase could be detected based on the power indices described above. 
The architecture of this algorithm is based on the qualitative understanding of human interaction data. The idea is to focus on the rising edges of the "hump" patterns which correspond to the initial intent expression phase. The peak location indicates the consensus attainment during the conflict. Furthermore, we noticed asynchronous actions among agents, implying different time frames for action phases detected in each participant. Our data collection involved the use of a beep signal to initiate participants' movement, ensuring that actions started only after the beep signal.

Figure \ref{fig:action_phase} depicts an example of interaction observation with 4 power signals, where raw peaks are first detected for each signal and then filtered by removing small outliers. We impose that possibility of changing an intent within the fastest observed human reaction time~\cite{murchison1934handbook} is zero. Thus, adjacent peaks are combined with a reaction time of $\delta T = 0.25$s. The rising moments of detected peaks are found by intersecting a horizontal line equal to a fraction ($\tau=0.1$) of the detected peak. The period for each signal is then determined to be the period from the rising moment to the peak location. Next, each period from all power signals is aggregated using a logical OR operator, with a "true" value corresponding within the period time frame, and "false" for outside of the period. The action phase is obtained by finding the corresponding time instance of the first boolean island. Where a boolean island is defined as a consecutive sequence of "true" values in a logical array. The algorithm is described in more detail in Algorithm \ref{alg:action_phase}.

Using multiple signals to determine the action phase improves robustness compared to relying on a single signal. Moreover, relying only on raw power (Equation \ref{eq:power}) is not reliable as the dot product might return zero when the force-velocity angle is 90$^\circ$, even though a participant might exert significant force to move the object.

It is worth noting that this algorithm is designed for offline data analysis, but it has critical implications for restricting regions of interest into compact time frames with a high likelihood of subjects acting on the same assigned goal. Ultimately, the action recognition problem is converted to supervised learning.




Please note that this version of the algorithm is designed to detect only the first round of actions. From our human studies, we observed that most negotiations are resolved after the first exchange of actions. However, in some cases, dyads may require more than one round of force exchange to resolve a conflict. For instance, in daily life situations, dyads may attempt to move in opposite directions by stretching a table, which results in a conflict that stops the interaction. In the next attempt, both dyads may change their goal and try to move in the reverse direction, leading to another compressive conflict. Only then, dyads may resort to language or assume a follower role to correctly determine the direction of the motion. Modifying the algorithm to detect subsequent force exchanges is part of our future work.

\section{Methodology}

\subsection{Data Preparation}

\subsubsection{Experiment}

A human study was designed to investigate the negotiation process in collaborative manipulation tasks within a challenging environment. A wooden tray ($2.1kg$) was used as a manipulated object. The experimental setup included three goals located 2.4m away from starting location, with a narrow angular separation of $40^\circ$. This angular configuration was selected to present a more complex scenario for the intent recognition task, as larger separation angles would simplify the task considerably. Prior to each trial, participants were provided with a goal configuration privately that specifies the location and the importance of the goal. The importance of the goal was parameterized through the addition of \textit{soft} and \textit{hard} goals. \textit{Soft} goals required participants to attempt to move toward their assigned direction, but they were allowed to concede and follow their partner's direction. On the other hand, \textit{hard} goals mandated that participants move towards their assigned direction and convince their partner if necessary. Additionally, some participants were assigned \textit{follower} role, with no specific goal except to follow their partner's direction. During the trials, participants were restricted to communicating through haptic channels, with verbal and other non-verbal communication prohibited. Participants were free to choose their preferred strategy to reach a consensus when a conflict arose. A detailed explanation of the experiment can be found in \cite{rysbek_robots_23}.

\begin{algorithm}[t]
\caption{Action Phase Detection}
\label{alg:action_phase}


Given $\tau<1$, $\delta T$\;
Given negotiation time frame $t \in [t_{neg}, t_{dec}]$, \\
Given power signals $P(t) = [|P_k|, P_k^1, ... , P_k^N], k \in \{1,2\}$\;

\ForAll{ $p$ in P}{
    $peaks$ = Find All Peaks in $p$\;
    Combine $peaks$ that are within $\delta T$ \;
    Remove outliers in $peaks$ with small magnitude\;
    $p_{dominant}$ is the first peak's magnitude\;
    $t_{dominant}$ is corresponding location in time\;
    \uIf{$p_{dominant}$ is $NULL$}{Continue\;}
    find $t_{start}<t_{dominant}$ when $p(t_{start}) = \tau p_{dominant}$\;
    store [$t_{start}$, $t_{dominant}$] in $periods$\;
}
$t_{bool}$ = $\bigcup periods$ in boolean vector form\;
[$t_0$, $t_f$] =  First boolean island in $t_{bool}$\;
\textbf{Return} Action Phase [$t_0$, $t_f$]\;
\end{algorithm}


Freedom in the choice of strategy by the dyads resulted in a considerable variety of behaviors in the interaction data. 
Broadly, participant actions (intent expression) fall into two categories: decisive and indecisive. Indecisive actions correspond to a subject trying to perceive the partner's intent, or probing actions that involve tentative movement towards a goal with the idea of giving up if the intent of the other participant is decisive. This is typical behavior for \textit{follower} and \textit{soft} goals. In contrast, decisive actions are unequivocally perceived by the other participant. When both participants choose a decisive action (with different goals) the result is typically a perceivable conflict where the stretching force magnitude exceeds 5N. In such interactions, the non-dominant partner exhibits \textit{opposing} behavior, characterized by negative applied power ($P_k<0$), which means the subject is not contributing to the motion, but being dragged away by the partner. Such interactions can result from both \textit{hard} and \textit{soft} goal configurations. We removed such opposing behavior and the participant actions that were unclear in intention due to weak signals when training the classifiers. However, we tested the classifiers on these actions and report the results in this work.




\subsubsection{Annotation}

The collected human interaction data is processed with the action phase detection algorithm described in section \ref{sec:first_mover}. The output of the algorithm is the action phase and associated power strength ($max(P_k(t))$ for $t \in [t_0, t_f]$) as depicted in Figure \ref{fig:training_scheme}. The period that precedes to action phase is considered an idle phase that starts at the beep signal generated during the experiment. By the design of the experiment, during the action phase subject attempts to go to the assigned goal direction. Therefore, data sampled from that action phase is the representation of the intent of the subject that was privately assigned to them. To account for the freedom of the strategy, particularly in \textit{soft} goal configuration where participants deviated from their assigned goal, the dataset is semi-automatically re-annotated to determine the accurate intent of the participant. Semi-automation is implemented according to interaction power and stretching force ($F_{stretch}=F_1-F_2$) cues. Annotators relied on the human body movements from the video when the behavior was ambiguous from the signals.

\subsubsection{Features}
\label{sec:features}

In the context of practical pHRI, it is crucial to utilize features that are readily available to the robot during runtime. In this study, we leverage force data at the grasp point, object position and velocity, linear acceleration, and angular velocity to infer the human partner's intent. It is unrealistic to assume that objects will be equipped with force-torque sensors in daily life. However, from the robot's force sensor, the applied force by the partner can be computed, given the inertia and acceleration of the object. In this work, we focus on inferring the participant's intent from its own applied force and velocities. A more realistic scenario would be to infer the collaborator's intent from the participant's own signals and actions. Although the signals of the collaborator can be computed from the participant's own data so the two problems are conceptually equivalent. Evaluating how this would work in practice is the subject of future research.

\begin{table}[t]
\centering
\begin{tabular}{lcccc} 
\hline \hline
\multirow{2}{*}{\textbf{Metric}}      & \multirow{2}{*}{\textbf{dim}} & \multicolumn{3}{c}{\textbf{Feature Set}}                         \\
                                      &                               & \textbf{\textbf{1}} & \textbf{\textbf{2}} & \textbf{\textbf{3}}  \\\hline 
Handle Velocity $v_k$                        &  2             &  \checkmark      &  -              &  -               \\
Applied Force $F_k$                          &  2             &  \checkmark      &  -              &  -               \\
Sum of the Forces $F_{sum}$                  &  2             &  \checkmark      &  -              &  -               \\
Stretch Force $F_{str}$                      &  2             &  \checkmark      &  -              &  -               \\
Raw Power $P_k$                              &  1             &  \checkmark      &  -              &  \checkmark      \\
Projected Velocity $v_k^i$                   &  N             &  \checkmark      &  \checkmark     &  \checkmark      \\
Projected Force $F_k^i$                      &  N             &  \checkmark      &  \checkmark     &  \checkmark      \\
Projected Sum of the Forces $F_{sum}^i$      &  N             &  \checkmark      &  \checkmark     &  -               \\
Projected Stretch Force $F_{str}^i$          &  N             &  \checkmark      &  \checkmark     &  -               \\
Projected Power $P_k^i$                      &  N             &  \checkmark      &  \checkmark     &  \checkmark      \\ \hline
\textbf{Number of unique signals}                &                &  24              &  15             &  10              \\
\textbf{Number of features with derivatives}              &                &  48              &  30             &  20              \\
\textbf{Number of features with statistics}               &                &  192             &  120            &  80              \\ \hline \hline
\end{tabular}
\caption{Principal signal content of each feature set. Dimensions are counted for $N=3$.}
\label{tab:feature_set_signals}
\vspace{-1.05cm}
\end{table}
%


To test how well the system adapts to different sets of available signals we evaluated it using three distinct feature sets, each designed to cater to a specific scenario. These feature sets are summarized in Table \ref{tab:feature_set_signals}. Feature Set 1 incorporates the participant's own applied force and a combination of the partner's force, as well as force and velocity measured in the spatial frame. Feature Set 2 comprises only scalar features, which are projections of force, velocity, and power. Feature Set 3 is the most conservative which includes only the participant's own scalar signals and disregards signals from the partner's side. An advantage of employing feature sets 2 and 3 is that they are independent of the spatial coordinates and contain only relative measurements with respect to the goal locations. This allows the model to generalize across different environments and coordinate systems, which is particularly relevant for pHRI applications where the robot interacts with objects in various contexts. Moreover, using only participant signals in set 3 eliminates the need for sensing the partner's input, which can simplify the hardware requirements and reduce the computational burden. We present a comprehensive discussion of the model's performance on each set in Section \ref{sec:results}.






\subsubsection{Neighborhood Sampling}
\label{sec:neiborhood_sampling}

We adopted a sliding window-based approach that is commonly used for time series data \cite{sana_unsup_representation_21, varol_multiclas_realtime_intent}. To achieve real-time intent recognition, we extract simple features from a time window of fixed length for signals provided in the feature set. The training data was generated by randomly sampling windows within the action phases identified using Algorithm \ref{alg:action_phase}. The minimum, maximum, mean, and standard deviations are extracted from each window. Each signal was acquired at 200Hz. 

Our neighborhood sampling is performed in two stages, which include uniform sampling across the action phase $[t_0, t_f]$ and half-normal distribution skewed to transition moments. The first stage helps to learn the general patterns of intent during the action phase while the second stage improved the accuracy in the transition periods. For idle cases, the half-normal distribution is right-skewed, while for actions, it is left-skewed. Using this approach, we observed a reduction in the transition delay of the real-time classifier by more than 0.1 seconds compared to uniform sampling only.

\subsection{Classifier Design}

\subsubsection{Support Vector Machine with ECOC}

Support Vector Machine (SVM)~\cite{cortes1995support} is a linear predictor that can train and classify nonlinear data points by first mapping the data set to a high-dimensional space using kernels. SVM then searches for a hyperplane that can separate the mapped training set with the largest possible margin into two binary classes. To extend SVMs for multiclass classification, we can use Error-Correcting Output Codes (ECOC)~\cite{dietterich1995ecoc}, which involves training multiple SVM learners based on coding and decoding designs. We chose the one-vs-all coding where each learner assigns one of the classes as a positive class ($+1$ code) while the remaining classes are treated as negative classes ($-1$ code). After training, when a sample is fed into the model, it generates a vector of scores with a length equal to the number of learners. The binary loss from each learner is calculated using this vector and the code matrix, and the one with the minimum aggregation of losses is selected as the predicted class. Based on \cite{madan_recognition}, we determined that multiclass SVMs with Gaussian kernels~\cite{hastie01statisticallearning}, were well-suited for our data. Using this approach, we trained four SVM learners where each corresponds to one of the four classes in our dataset. 


\subsubsection{Ensemble Learning with Decision Trees}
Inspired by \cite{alsaadi_resolving_conflicts}, we chose random forests~\cite{breiman2001random} as one of our classifiers. The random forest algorithm generates a bootstrap sample from the entire training set and uses it to train a classification and regression tree (CART)~\cite{breiman2017classification} based on the values in an independent and identically distributed random vector. During testing, the output is determined by selecting the class with the highest number of votes from the predictions of the trees for the given input.

Additionally, we explored another popular ensemble method known as Adaptive Boosting (AdaBoost) for multiclass classification~\cite{freund1997decision}. AdaBoost first distributes equal weights to all training samples and then trains a decision stump (decision trees with one split). The weights of misclassified observations increase while the weights of correctly classified ones decrease. It repeats the process for certain iterations and a more complicated error metric known as pseudo-loss~\cite{freund1997decision} is used. The prediction method is similar to random forests except that the votes are weighted for each learner based on the pseudo-loss.


\subsubsection{Fully connected Neural Network}




Feedforward neural networks (FNNs), also known as multilayer perceptrons (MLPs), are a classic method for classifying nonlinear data~\cite{Goodfellow-et-al-2016}. In this study, we trained a neural network with a specific structure, consisting of an input layer ($n_{0}$) sized according to the number of features, followed by two fully connected layers ($n_{1}, n_{2}$) where the number of neurons in each layer is proportional to the previous layer's size ($n_{1}=\alpha_1 n_{0}, n_{2}=\alpha_2 n_{1}, 2/3 \leq \alpha_1, \alpha_2 \leq 2$). Specific values of the $\alpha_1$ and $\alpha_2$ were determined with Bayesian hyper-parameter optimization method ~\cite{snoek2012practical}. The motivation for choosing this specific layer structure was to balance the data-to-model size ratio. With the above conservative structure of the network, the total number of training weights grows quadratically as a function of the number of input neurons:
\begin{equation}
    N = n_{0}^2 (\alpha_1+\alpha_1^2\alpha_2) + \alpha_1\alpha_2 n_{0}n_{f}
\end{equation}
For example, with a maximum of 192 input features, adequate neural network training would require a training set of at least 37,000 sampled windows, which is not feasible with our current dataset. For comparison, our dataset size is $\approx 13,000$. The final layer ($n_{f}$) has $N+1$ outputs that correspond to the number of classes. We used the rectified linear unit (ReLU) as the activation function for the first two layers and the Softmax function for the final layer. To minimize the cross-entropy loss, the limited-memory Broyden-Fletcher-Goldfarb-Shanno quasi-Newton algorithm (LBFGS) was employed ~\cite{nocedal2006numerical}.


\subsection{Evaluation}



A total number of interactions (619) are divided into testing and training sets each consisting of 15\% and 85\% respectively. Then, each dyad's actions in the training set were sampled with the neighborhood sampling strategy described in Section \ref{sec:neiborhood_sampling}. To avoid overfitting, 5-fold cross-validation was used by each classifier. Models were tested on the similarly sampled data from test set interactions. As an overall accuracy metric, we report a confusion matrix, as well as precision, recall, and F1 scores, as they are suitable for imbalanced class distribution. For compact accuracy, the metric macro-averaged F1 score is used throughout this study.

Another way to assess the model performance is through transition accuracy at the signal level. When used in pHRI applications, the classifier's output is often post-processed using a majority-voting technique \cite{varol_multiclas_realtime_intent}. While this introduces fixed time delays depending on voting buffer size, it enhances the accuracy and reduces short-term false transitions. The most objective evaluation method, in this case, is comparing true instantaneous human action with classifier predictions. However, only high-level participant goal configurations that are static during the action phase are available in this study. Comparing with assigned goals is unsuitable since agents often express intents in multiple directions within a single action phase. To accommodate this, we evaluate the model based on two additional metrics, \textit{Successful Transition Rate} and \textit{Negotiated Goal Prediction}. \textit{Successful Transition Rate} is defined as a classifier predicting an assigned goal during the action phase. \textit{Negotiated Goal Prediction} is defined as predicting the final goal of the dyads. It is evaluated only for instances where a participant \textit{opposed} to one direction and then changed the intent. Finally, we note that signal level accuracy was performed on \textit{opposing} instances that were not part of the training set. This provides evidence for the generalization of the intent recognition model.

\section{Results and Discussion}
\label{sec:results}

\begin{table}[t]
\resizebox{!}{\height}{%
\begin{tabular}{lccc}
\hline
\hline
\multirow{2}{*}{Models} & \multicolumn{3}{c}{Feature Set} \\
                  & 1                & 2                & 3                 \\ \hline
\textbf{SVM}      & 0.7552           & 0.7292           & 0.6583            \\ 
\textbf{AdaBoost} & \textbf{0.7660}  & \textbf{0.7482}  & \textbf{0.7279}   \\ 
\textbf{NN}       & 0.7591           & 0.7331           & 0.7038            \\ \hline \hline
\end{tabular}%
}
\caption{Macro-average F1 score of the goal classes from each model with different feature sets. Window size is fixed at 60.}
\label{tab:featureset}
\vspace{-0.8cm}
\end{table}

\begin{table}
\resizebox{\columnwidth}{!}{%
\begin{tabular}{llcccc}
\hline \hline
\multirow{2}{*}{Models}   & \multirow{2}{*}{Evaluation metrics} & \multicolumn{4}{c}{Class}         \\
                          &                                     & Idle   & Goal 1 & Goal 2 & Goal 3 \\ \hline
\multirow{3}{*}{\textbf{SVM}}      & Precision                           & 0.9892 & 0.8502 & 0.6046 & 0.8227 \\
                          & Recall                              & 0.9828 & 0.7517 & 0.7008 & 0.8146 \\
                          & F1 score                            & 0.9860 & 0.7979 & 0.6491 & 0.8186 \\ \hline
\multirow{3}{*}{\textbf{AdaBoost}} & Precision                           & 0.9663 & 0.8489 & 0.6516 & 0.8134 \\
                          & Recall                              & 0.9850 & 0.7815 & 0.7083 & 0.7976 \\
                          & F1 score                            & 0.9756 & 0.8138 & 0.6788 & 0.8054 \\ \hline
\multirow{3}{*}{\textbf{NN}}       & Precision                           & 0.9807 & 0.8363 & 0.6122 & 0.8377 \\
                          & Recall                              & 0.9828 & 0.7782 & 0.7235 & 0.7805 \\
                          & F1 score                            & 0.9818 & 0.8062 & 0.6632 & 0.8081 \\ \hline \hline
\end{tabular}%
}
\caption{Precision, Recall, and F1 scores of each model measured on the test set. The inference is performed with Feature Set 1 and the window size of 60.}
\label{tab:evalmetrics}
\vspace{-0.9cm}
\end{table}

\begin{figure}[t]
    \centering
    \resizebox{\columnwidth}{!}{\includegraphics[trim={4.5cm 7cm 4.5cm 6.3cm}]{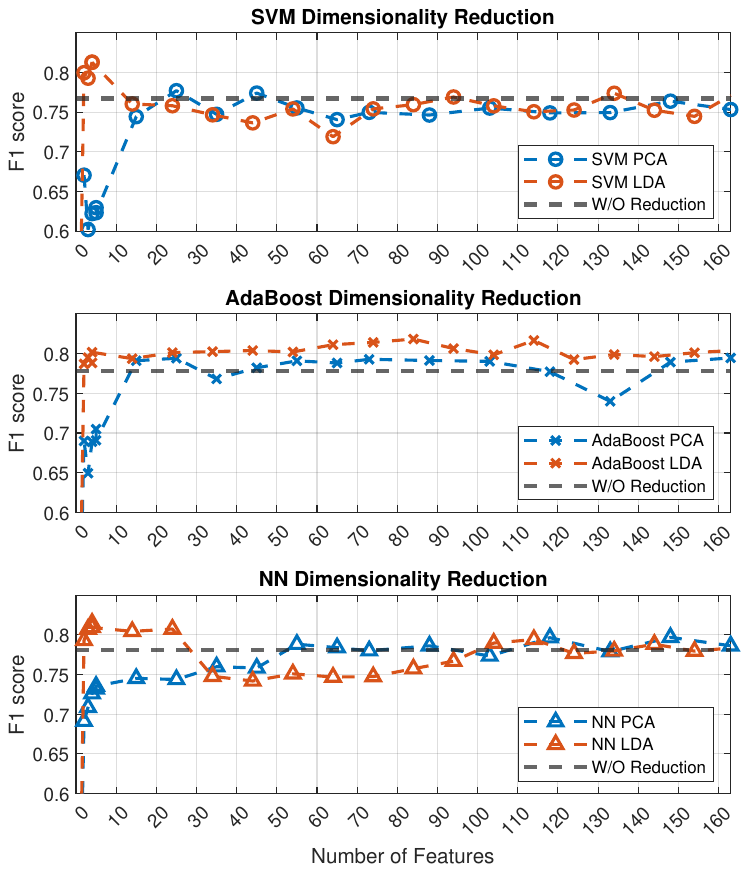}}
    \caption{Results of Dimensionality Reduction applied on each model using Feature Set 1.}
    \label{fig:pca_lda}
    \vspace{-0.1cm}
\end{figure}

We used two common dimensionality reduction techniques prior to classifier training: Principal Component Analysis (PCA) and Linear Discriminant Analysis (LDA). PCA is an unsupervised technique that linearly transforms the data into a new coordinate system that maximizes the variance of the data by preserving the maximum information \cite{jolliffe2005principal}. LDA, on the other hand, is a supervised technique that transforms the data into a coordinate system that maximizes the separation of classes by taking into account class information~\cite{fisher1936use}. As shown in Figure~\ref{fig:pca_lda}, both PCA and LDA improved model performance compared to no reduction. LDA consistently outperformed PCA and required a smaller number of features to achieve optimal performance, reaching its peak faster than PCA. Although AdaBoost had slightly better accuracy results for 84 top principal components, SVM and FNN required only 4 principal components to achieve optimal performance. Additionally, a smaller number of features reduced the complexity of the FNN architecture and improved the training time performance for both SVM and FNN. In the rest of the discussions, LDA dimensionality reduction is used for all models.

\begin{table}[t]
\centering
\resizebox{!}{\height}{%
\begin{tabular}{lcccc}
\hline
\hline
\multirow{2}{*}{Models} & \multicolumn{4}{c}{Window Size}   \\
                        & 20              & 40              & 60              & 80              \\ \hline
\textbf{SVM}            & 0.6985          & \textbf{0.7383} & 0.7552          & 0.7525          \\
\textbf{AdaBoost}       & \textbf{0.7054} & 0.7380          & \textbf{0.7660} & \textbf{0.7796} \\
\textbf{NN}             & 0.6959          & 0.7343          & 0.7591          & 0.7504          \\\hline
\textbf{Delay(AdaBoost)}& 0.231s          & 0.246s          & 0.265s          & 0.350s          \\\hline\hline
\end{tabular}%
}
\caption{Macro-average F1 score of the goal classes from each model with different window sizes. The inference is performed with Feature Set 1.}
\label{tab:windowsize}
\vspace{-0.8cm}
\end{table}

\begin{table}
\resizebox{\columnwidth}{!}{%
\begin{tabular}{lccc}
\hline
\hline
   Metrics                                   & \textbf{SVM}    & \textbf{AdaBoost} & \textbf{NN}     \\ \hline
Average Precision of goal classes            & 0.7591 & 0.7713   & 0.7621 \\
Successful Transition Rate                   & 0.9361 & 0.9361   & 0.9535 \\
Negotiated Goal Prediction                   & 0.9630 & 0.9630   & 1.0000 \\ 
\hline
\hline
\end{tabular}%
}
\caption{Signal Level Accuracy}
\label{tab:signal_level_accuracy}
\vspace{-1cm}
\end{table}

Understanding the trade-off between model accuracy and available signals is crucial in practical applications. Moreover, training a classifier with force and velocity measured in a specific spatial frame may lead to overfitting and require re-training for a new environment. To address this issue, we evaluated the model with three different feature sets. The macro-averaged F1 score of goals for each model is summarized in Table \ref{tab:featureset}. Results indicate that the model's performance decreases as the number of available signals reduces. However, Feature Set 2 and 3 perform competitively compared to Feature Set 1. Interestingly, Feature Set 3 had a significant impact on the performance of SVM, resulting in a drop of around 0.1. This analysis confirms that coordinate-invariant conservative features can be used with reasonably good performance, by trading a small drop in F1 score ($\approx2\%$).

\begin{figure}
    \centering
    \includegraphics[width=1.1\columnwidth, trim={2cm 6.5cm 0.5cm 6.5cm}]{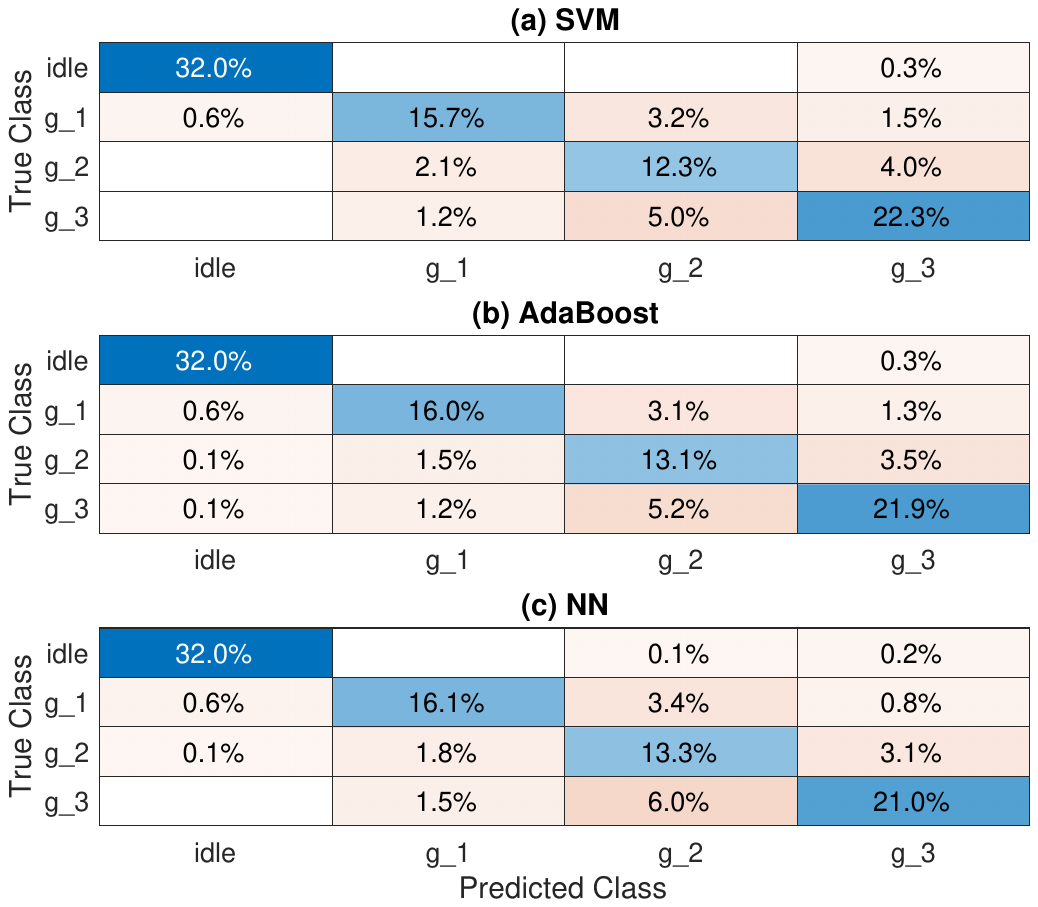}
    \caption{Total normalized confusion matrix for each model performed with Feature Set 1 and Window Size 60 (blank represents $0\%$).}
    \label{fig:confusionchart}
    \vspace{-0.5cm}
\end{figure}

Overall model performance is presented in the confusion matrix in Figure \ref{fig:confusionchart}. \textit{Idle} class is well-separated compared to the other classes. This is expected because during this phase no participant exerts force and the object stays still which is easy to distinguish. However, $g_2$ is the most ambiguous to classify compared to $g_1$ and $g_3$ for the following reasons. First, goal 2 is geometrically close to both 1 and 3. Second, participants are oriented towards $g_2$ at the beginning of the interaction. Hence, the initial phase of actions inevitably points towards $g_2$ and the misclassification rate increases even though the model correctly predicts it. Third, the proportion of delay and the total action phase is relatively high, although the transition delay is short ($\approx0.25s$). Hence, it results in a lower accuracy rate of goal 2 predictions. As each delay period is counted as a misclassification, it comprises a significant portion of the negotiation phase, which might be the reason for the F1 scores for goal $g_2$ being bounded at $\sim80\%$. Finally, $g_1$ and $g_3$ had a high angular separation of $80^\circ$, thus model confusion for these cases is relatively low. Precision, recall, and F1 scores are presented in Table \ref{tab:evalmetrics}.

Choosing an appropriate window size is crucial in intent recognition, as it affects both the delay and accuracy of the model. A smaller window size reduces the delay but sacrifices accuracy, while a larger one enhances accuracy at the expense of delay. Additionally, the window size must be limited by the reaction time of the average human, so that the data used for prediction corresponds to a single action. In this study, the optimal window size was found to be 60 data points, corresponding to a duration of 0.3 seconds. AdaBoost performed slightly better than the other models for this window size. Furthermore, models trained on Feature Set 2 showed similar results to those trained on Feature Set 1 for different window sizes. Details of this analysis are shown in Table \ref{tab:windowsize}.

Finally, post-voting model performance was evaluated based on the successful transition rate. As can be seen from Table \ref{tab:signal_level_accuracy}, all three models show similarly high accuracy ($~94\%$). However, when further analyzed for the incorrect cases, out of 11 misclassified cases 8 were due to mislabelling, 2 were because of force angle distortion due to strong conflict (in such cases, dyads cannot identify opposer's intended direction, but perceives opposition only). The only failed instance was due to the expression of weak intent, where none of the classifiers were sensitive enough. While the NN had no failed instances, AdaBoost was chosen due to its slightly superior overall performance.

\begin{figure*}
     \centering
     \begin{subfigure}[b]{0.8\textwidth}
         \centering
         \includegraphics[width=0.8\textwidth, trim={5.5cm 0cm 6cm 0cm}]{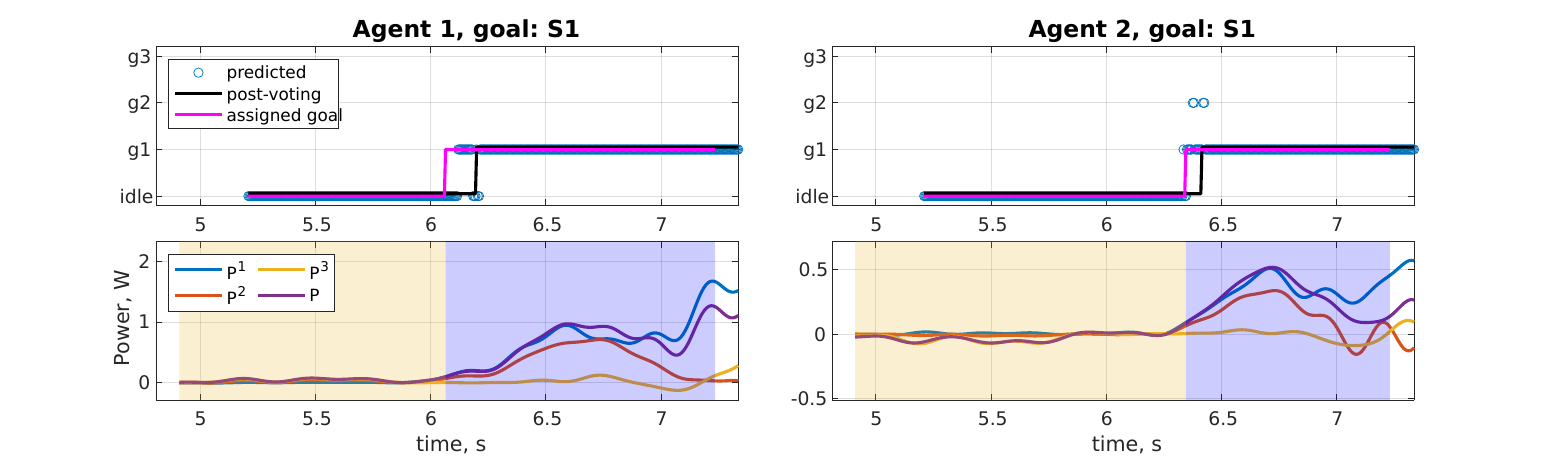}
         \caption{Non-conflciting Interaction. Dyads negotiated to $g_1$.}
         \label{fig:voting_4}
     \end{subfigure}
     \begin{subfigure}[b]{0.8\textwidth}
         \centering
         \includegraphics[width=0.8\textwidth, trim={5.5cm 0cm 6cm 0cm}]{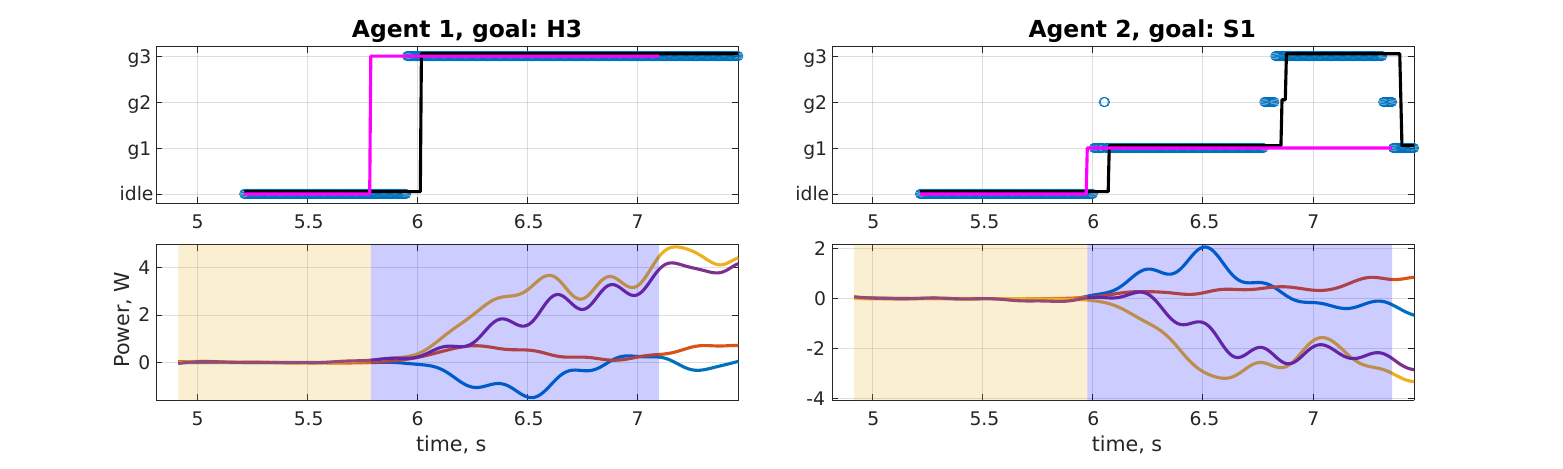}
         \caption{Conflicting interaction. Dyads negotiated to $g_3$. Note that, the classifier correctly recognizes both transitions for participant 2: the first intention to $g_1$, then $g_3$ as a result of negotiation. }
         \label{fig:voting_11}
     \end{subfigure}
     \begin{subfigure}[b]{0.8\textwidth}
         \centering
         \includegraphics[width=0.8\textwidth, trim={5.5cm 0cm 6cm 0cm}]{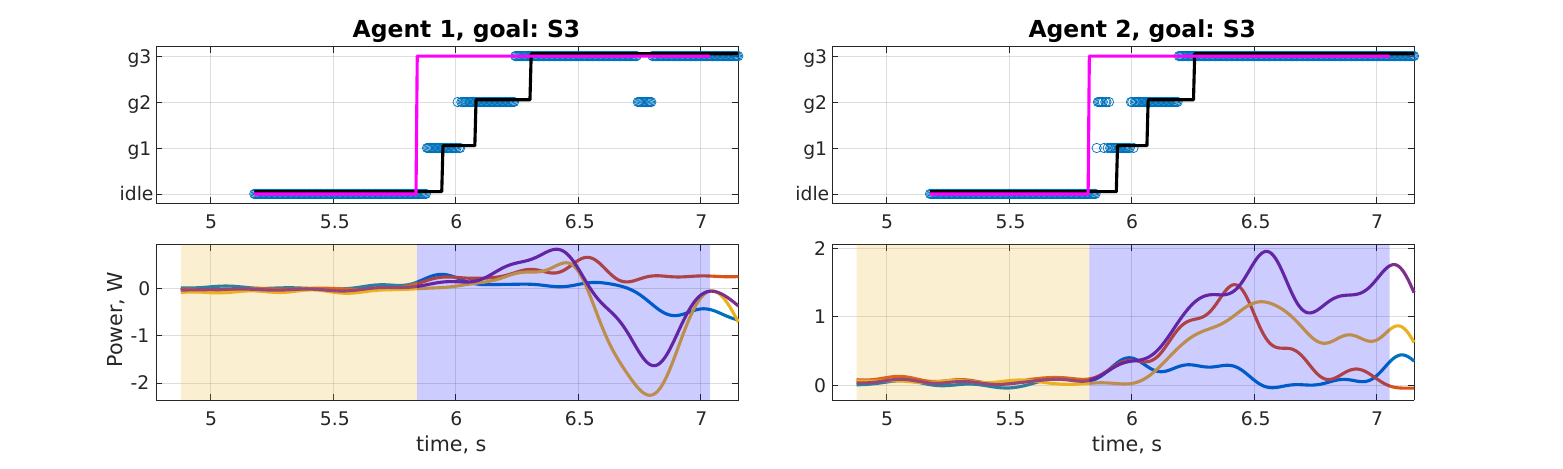}
         \caption{An example where both dyads were indecisive. Dyads negotiated to $g_3$.}
         \label{fig:voting_17}
     \end{subfigure}
     \caption{Model performance on different scenarios of interaction. In each subplot, the first row shows real-time classifier prediction for participants 1 (left column) and 2 (right column) in a dyad. In the second row, the power metrics for each participant are drawn. Yellow shaded area corresponds to the \textit{idle} and blue area for the action phase. The goals of the participants are indicated in the subfigure titles. In this context, \textit{H} and \textit{S} stand for \textit{hard} and \textit{soft}, respectively, while the numbers correspond to the goal locations. The signals are tested with AdaBoost, trained under Feature Set 1, Window Size 60. A voting Buffer of 25 was applied in all of the examples.}
     \label{fig:voting_shcemes}
     \vspace{-0.5cm}
\end{figure*}

The accuracy of the negotiated goal prediction was evaluated on \textit{opposing} instances. In all 33 actions, the models achieved a high classification rate, with the NN having zero errors. This suggests that the classifier is capable of accurately predicting the final negotiated goal even in scenarios where participants change their initial intentions. The example given in Figure \ref{fig:voting_shcemes}(b) shows how the AdaBoost classifier recognizes both participants' initial intent towards $g_3$ and $g_1$, respectively, which caused the conflict. After that, participant 2 gives up their own goal and agrees to go to $g_3$, and the model correctly transitioned to the final negotiated goal 3. This highlights the importance of using an objective evaluation method for intent recognition, as relying solely on assigned goal configuration may not provide an accurate representation of the participants' intentions.

%








\section{Conclusion}

%
%
%


This paper introduces a real-time capable human intent recognizer designed for collaborative manipulation tasks. The recognizer predicts the partner's action using force and velocity signals experienced by the participant. Moreover, our study delves into force signal communication, an underexplored modality in pHRI. 


An important contribution of our work is a procedure for generating a training dataset given that the real human intent in highly dynamic tasks such as collaborative manipulation is next to impossible to establish. Central to the dataset is a carefully designed human-human collaborative manipulation study where each participant is assigned private goals. Once a training dataset has been established, novel features such as power projected to a particular goal direction were used to train several classifiers for predicting the intent. An extensive analysis of different classifiers has been performed, using various metrics that reflect both naive classifier performance as well as performance in the realistic pHRI application. The best-performing classifier showed excellent generalization performance, with excellent results on interactions that were not part of the training set.

However, the current classifier has some limitations. It lacks the ability to evaluate the strength of intent, which is an important factor in determining whether the robot should lead or follow, and thus how it should act. Though we provide theoretical arguments that frame invariant features can generalize to different environments, this has to be tested to different goal numbers and locations. Moreover, the current training set was based on only the first round of force exchanges after the idle state, while in reality, multiple rounds of actions occur in an ad-hoc manner after the first interaction.


To address these limitations, we aim to devise a universal model that can generalize to arbitrary numbers and locations of goals, as well as multiple rounds of force exchanges. Achieving this eliminates the need for training and facilitates the practical use of such robots in daily life, where the destination of co-manipulation tasks constantly change depending on the object type and task requirements. Another extension of this work is using this intent recognizer for real-time feedback control of pHRI.

\begin{acks}
This work has been supported by the National Science Foundation grants IIS-1705058, CMMI-1762924, and CCF-2240532.
\end{acks}



\bibliographystyle{ACM-Reference-Format}
\bibliography{icmi_2023_arxiv.bib}


\end{document}